\documentclass{article}

\newif\ifPrintAppendix
\PrintAppendixtrue
\newcommand{\WithOrWithoutAppendix}[2]{\ifPrintAppendix#1\else#2\fi}

\usepackage[final]{neurips_data_2021}

\usepackage[utf8]{inputenc}
\usepackage[T1]{fontenc}   
\usepackage{hyperref}      
\usepackage{url}           
\usepackage{booktabs}      
\usepackage{amsfonts}      
\usepackage{nicefrac}      
\usepackage{microtype}     
\usepackage{xcolor}        

\usepackage[english]{babel}
\usepackage{natbib}

\usepackage{adjustbox}
\usepackage{siunitx}
\usepackage{listings}
\usepackage{caption}
\usepackage{subcaption}
\usepackage{todonotes}
\usepackage{textcomp}
\makeatletter
\presetkeys
    {todonotes}
    {inline}{}
\makeatother
\usepackage[shortlabels,inline]{enumitem}
\newcommand\hide[1]{}
\newcommand{\numdatasets}{72}

\definecolor{mygreen}{rgb}{0,0.6,0}
\definecolor{mygray}{rgb}{0.5,0.5,0.5}
\definecolor{mymauve}{rgb}{0.58,0,0.82}
\title{OpenML Benchmarking Suites}

\author{
  Bernd Bischl$^1$\thanks{Authors are ordered alphabetically. Correspondence to \{bernd.bischl | giuseppe.casalicchio\}@lmu.de.}~, Giuseppe Casalicchio$^1$, Matthias Feurer$^2$, Pieter Gijsbers$^3$, Frank Hutter$^{2,4}$, \\
  \textbf{Michel Lang$^5$, Rafael G. Mantovani$^6$, Jan N. van Rijn$^7$, Joaquin Vanschoren$^{3}$} \\
  $^1$ Department of Statistics, LMU Munich, Germany \\
  $^2$ Department of Computer Science, University of Freiburg, Germany \\
  $^3$ Department of Computer Science, Eindhoven University of Technology, the Netherlands\\
  $^4$ Bosch Center for Artificial Intelligence \\
  $^5$ Department of Statistics, TU Dortmund University, Germany \\
  $^6$ Federal Technology University Paran\'a (UTFPR), Brazil\\
  $^7$ Leiden Institute of Advanced Computer Science (LIACS), Leiden University, the Netherlands\\
}

\hypersetup{pdftitle={OpenML Benchmarking Suites},pdfauthor={Bernd Bischl, Giuseppe Casalicchio, Matthias Feurer, Frank Hutter, Michel Lang, Rafael G. Mantovani, Jan N. van Rijn, Joaquin Vanschoren}}

\begin{document}

\maketitle
\setcounter{footnote}{0}

\begin{abstract}
Machine learning research depends on objectively interpretable, comparable, and reproducible algorithm benchmarks. 
We advocate the use of curated, comprehensive \textit{suites} of machine learning tasks to standardize the setup, execution, and reporting of benchmarks. We enable this through software tools that help to create and leverage these benchmarking suites. 
These are seamlessly integrated into the OpenML platform, and accessible through interfaces in Python, Java, and R.
OpenML benchmarking suites 
\begin{enumerate*}[(a),noitemsep]
	\item are easy to use through standardized data formats, APIs, and client libraries; 
    \item come with extensive meta-information on the included datasets; and
    \item allow benchmarks to be shared and reused in future studies.
\end{enumerate*}
We then present a first, carefully curated and practical benchmarking suite for classification: the \textbf{OpenML} \textbf{C}urated \textbf{C}lassification benchmarking suite 20\textbf{18} (OpenML-CC18).
Finally, we discuss use cases and applications which demonstrate the usefulness of OpenML benchmarking suites and the OpenML-CC18 in particular.
\end{abstract}

\section{Introduction}
\label{sec:intro}

Algorithm benchmarks shine a beacon for machine learning research. They allow us, as a community, to track progress over time, identify challenging issues, to raise the bar and learn how to do better. To learn as much as possible from them, they must include well-designed, challenging sets of tasks, be easily accessible and practical to use. Evaluations of algorithms on these tasks should be performed in standardized ways to support a rigorous analysis and clear conclusions. And above all, these evaluations must be easy to find, easily interpretable, reproducible, and directly comparable to evaluations run by other scientists.

The OpenML platform \citep{OpenML2013} already serves thousands of datasets  together with tasks in a machine-readable way. Tasks define the evaluation procedure for a specific dataset. Concretely, a task contains a reference to a dataset, information on the task type (e.g., classification or regression), the target feature (in the case of supervised problems), the evaluation procedure (e.g., k-fold CV, hold-out),  the specific splits for that procedure, and the target performance metric, which together allow for reproducible evaluation schemes. OpenML is also integrated into many machine learning libraries, so that fine details about machine learning models (or pipelines) and their performance evaluations can be automatically collected. This integration allows experiments to be automatically shared and organized on the platform, linked to the underlying datasets and tasks.
However, OpenML did not yet facilitate the simple creation and sharing of well-designed benchmark suites and results of experiments ran on them.

We introduce a novel benchmarking layer on top of OpenML, fully integrated into the platform and its APIs, that streamlines the creation of \textit{benchmarking suites}, i.e., collections of tasks designed to thoroughly evaluate algorithms. These suites can then be easily imported, used in systematic benchmarking experiments, and the results can be automatically shared and organized on the OpenML platform, where they can be easily searched, reused and compared to the results of others.
We develop tools that allow for creating a well-defined benchmark suite, and propose a new benchmark suite designed with these tools: the \textbf{C}urated \textbf{C}lassification benchmarking suite 20\textbf{18} (OpenML-CC18).

In short, the contributions of this paper are as follows: 
\begin{enumerate*}[(1),noitemsep]
    \item we advocate the use of curated, comprehensive \textit{suites} of machine learning tasks (i.e., a dataset with meta-information about the evaluation procedure) to standardize benchmarking, 
    \item we provide software tools to easily create and use these benchmarking suites, 
    \item we propose a new benchmark suite (OpenML-CC18), 
    \item have a closer look at an existing AutoML benchmark suite, and
    \item discuss their impact on machine learning research.
\end{enumerate*}
\footnote{\label{footnote-arxiv}We previously published a preprint on arXiv, which has already been used in new research. This is the reason we can both introduce OpenML-CC18 and benchmark suites technology, but also review their use. For example, the AutoML benchmark suite was created with the technology described in this paper (and the preprint).}

We will first discuss related work. Next, we explain how OpenML benchmarking suites work and how to use them in practice. We then present the OpenML-CC18 and review other benchmarking suites, including the AutoML benchmark. Finally, we discuss the impact of benchmarking suites on machine learning research and present our conclusions.  

\section{A Brief History of Benchmarking Suites}
\label{sec:history}

The machine learning field has long recognized the importance of dataset repositories. The UCI repository~\citep{UCI:2017} and LIBSVM~\citep{chang2011libsvm} offer a wide range of datasets. Many more focused repositories also exist, such as UCR~\citep{UCRArchive} for time series data and Mulan~\citep{mulan} for multilabel datasets. Some repositories also provide programmatic access. \texttt{\href{https://www.kaggle.com/}{Kaggle.com}} and PMLB~\citep{pmlb} offer a Python API for downloading datasets, skdata~\citep{bergstra2015skdata} offers a Python API for downloading computer vision and natural language processing datasets, and KEEL~\citep{KEEL} offers a Java and R API for imbalanced classification and datasets with missing values.

Several platforms can also link datasets to reproducible experiments (similar to OpenML tasks). Reinforcement learning environments such as the OpenAI Gym~\citep{openai_gym} run and evaluate reinforcement learning experiments, the COCO suite standardizes benchmarking for blackbox optimization~\citep{hansen-oms20} and ASLib provides a benchmarking protocol for algorithm selection~\citep{bischl-aij16a}. The Ludwig Benchmarking Toolkit orchestrates the use of datasets, tasks and models for personalized benchmarking and so far integrates the Ludwig deep learning toolbox~\citep{narayan-neurips2021a}. \texttt{\href{https://paperswithcode.com/}{PapersWithCode}} maintains a manually updated overview of model evaluations linked to datasets.

Although for many years machine learning researchers have benchmarked their algorithms on some subset of these datasets, this has not yet led to standardized benchmarks that can be easily compared between individual studies. This often results in suboptimal shortcuts in study design, producing rather small-scale experiments that should be interpreted with caution \citep{Aha:1992p455}, are hard to reproduce \citep{Pedersen:2008p12980,Hutson2018}, and even lead to contradictory results \citep{Keogh:2003p4930}. An often criticized aspect is the competitive mindset in benchmarking which focuses too much on dominating the state-of-art on a few datasets, instead of a rigorous and informative analysis of large-scale studies, including negative results where popular algorithms fail \citep{sculley2018winner}.

\section{OpenML}
OpenML is a collaborative platform that allows anyone to share new datasets, and enables anyone to easily import these datasets and subsequently share their own models and experiments run on them. It organizes everything based on four fundamental, machine-readable building blocks: (1) the \textit{data}, (2) the machine learning \textit{task} to be solved, specifying the dataset, the task type (e.g., classification or regression), the target feature (in the case of supervised problems), the evaluation procedure (e.g., k-fold CV, hold-out), the specific splits for that procedure, and the target performance metric (3) the \textit{flow} which specifies a machine learning pipeline that solves the \textit{task}, and (4) the \textit{run} that contains experiment results (e.g., predictions and performance evaluations) when a \textit{flow} is executed on a \textit{task} (see \citet{OpenML2013} for more details). OpenML goes beyond the platforms mentioned in Section~\ref{sec:history}, as it includes extensive programmatic access to all datasets, tasks, flows, and runs, comprehensive logging of experiments, and automated sharing of results, which have enabled the collection of millions of publicly shared and reproducible experiments, linked to the exact datasets, machine learning pipelines and hyperparameter settings. OpenML offers bindings with the Java, Python and R ecosystems \citep{Rijn2016, feurer-jmlr2021a, Casalicchio2017} to provide easy integration in common machine learning tools, workflows, and environments. An introduction and detailed information can be found on \url{https://docs.openml.org}.

\begin{figure}[t]
\includegraphics[width=\linewidth]{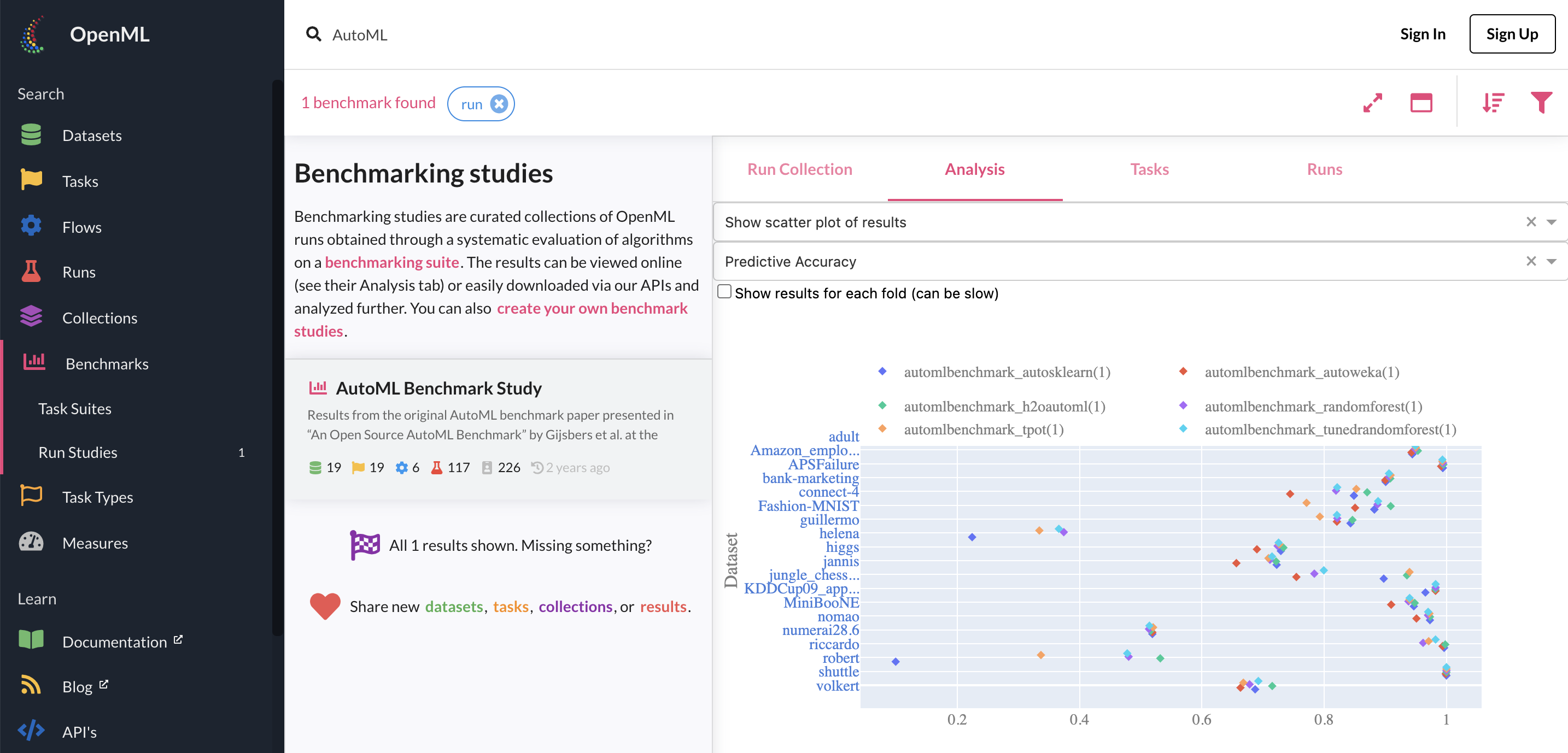}
\caption{OpenML website showing a list of benchmark studies on the left, and interactive exploration of the results of the AutoML Benchmark (see Section \ref{automlbenchmark}) on the right. Can be viewed online at \url{https://www.openml.org/s/226}.}
\centering
\label{website}
\end{figure}

\section{OpenML Benchmarking Suites}
\label{omlsuite}

As with any platform where people can upload new datasets, an overwhelming amount and variety of datasets is available, and it can be unclear how well they are curated. We designed OpenML benchmarking suites as a remedy to allow researchers to compile and publish well-defined collections of curated tasks and datasets, and collect benchmarking results from many scientists in a single place. More precisely, we define:

\emph{An OpenML benchmarking suite is a set of OpenML tasks carefully selected to evaluate algorithms under a precise set of conditions.}

Using a set of \textit{tasks} instead of a set of \textit{datasets} makes experiments performed on them comparable and reproducible. Compared to other (static) collections of datasets, the use of OpenML benchmarking suites has the following advantages:
\begin{itemize}[noitemsep,leftmargin=*]
    \item Easy creation of benchmarks (see Section~\ref{createsuites}): OpenML hosts thousands of datasets, and scientists can easily filter them down to those needed for their benchmarks (see Sections~\ref{cc18} and~\ref{sec:further_suites} for examples). 
    \item Convenient access and sharing of suites: Each suite receives a unique ID, which can be used to retrieve the suite via APIs, and via its own webpage. Figure \ref{website} illustrates how results collected on these suites can be explored online.
    \item Permanence and provenance: Because benchmarking suites are its own entity on OpenML, it is clear who created them (provenance). It also guarantees no one but the original creator can edit or remove the suite (permanence), this is an advantage over the previously used community tagging mechanism which allowed any user to add tasks to a suite.
    \item Community of practice: Curated benchmark suites allow scientists to thoroughly benchmark their machine learning methods without having to worry about finding and selecting datasets for their benchmarks.
    \item Building on existing suites: Scientists can extend, subset, or adapt existing benchmarking suites to correct issues, raise the bar, or run personalized benchmarks.
    \item Reproducibility of benchmarks: Based on machine-readable OpenML \textit{tasks}, with detailed instructions for evaluation procedures and train-test splits, shared results are comparable and reproducible.
    \item Conducting benchmark studies: After creating an OpenML benchmarking suite, existing and new experiments (\textit{runs}) on the underlying \textit{tasks} can be associated with the suite. This is also illustrated in Figure~\ref{joyplot}. Such data reuse bootstraps the creation of new benchmark studies that can analyze existing machine learning algorithms in new ways, or to design new challenging benchmark suites.
    \item Collaborative work: OpenML benchmarking suites benefit from the OpenML community, where users can help to identify and report bugs and errors in the contained datasets.
    \item Dynamic benchmarks: Benchmarks are never perfect, and when used for a long time, scientists may overfit on specific sets of tasks. However, benchmarking suites can be easily corrected and extended over time (e.g., on a yearly basis), leading to dynamic benchmarks that respond to novel concerns, and evaluate methods on new and ever more challenging tasks. More than providing a snapshot, this allows longitudinal studies that truly track progress over time.
\end{itemize}

\begin{figure}[!hb]
  \begin{center}
    \begin{subfigure}{\textwidth}
\begin{lstlisting}[language=Python]
from openml import config, study, tasks, runs, extensions
from sklearn import compose, impute, metrics, pipeline, preprocessing, tree

clf = pipeline.make_pipeline(
    compose.make_column_transformer(
        (impute.SimpleImputer(), extensions.sklearn.cont),
        (preprocessing.OneHotEncoder(handle_unknown='ignore'), extensions.sklearn.cat),
    ),
    tree.DecisionTreeClassifier(max_depth=1)
)  # build a fast and simple classification pipeline

benchmark_suite = study.get_suite('OpenML-CC18')       # obtain the benchmark suite
# config.apikey = 'FILL_IN_OPENML_API_KEY'             # uploading to OpenML requires an API key

run_ids = []
for task_id in benchmark_suite.tasks:                  # iterate over all tasks
    task = tasks.get_task(task_id)                     # download the OpenML task
    X, y = task.get_X_and_y()                          # get the data (not used in this example)
    run = runs.run_model_on_task(clf, task)            # run classifier on splits given by the task
    score = run.get_metric_fn(metrics.accuracy_score)  # compute and print the accuracy score
    print(f'Data set: {task.get_dataset().name}; Accuracy: {score.mean():.2}')
    run.publish()
    run_ids.append(run.id)

benchmark_study = study.create_study(                  # create a study to share the set of results
    name="CC18-Example",
    description="An example study reporting results of a decision stump.",
    run_ids=run_ids,
    benchmark_suite=benchmark_suite.id
)
benchmark_study.publish()
print(f"Results are stored at {benchmark_study.openml_url}")
\end{lstlisting}
      \caption{Python, available as pypi package  \href{https://pypi.org/project/openml/}{OpenML}
      }
    \end{subfigure} 
    \begin{subfigure}{\textwidth}
\begin{lstlisting}[language=Java]
public static void runTasksAndUpload() throws Exception {
  OpenmlConnector openml = new OpenmlConnector("FILL_IN_OPENML_API_KEY");
  Study benchmarksuite = openml.studyGet("OpenML-CC18", "tasks");           // obtain the benchmark suite
  Classifier tree = new REPTree();                                          // build a Weka classifier
  for (Integer taskId : benchmarksuite.getTasks()) {                        // iterate over all tasks
    Task t = openml.taskGet(taskId);                                        // download the OpenML task
    Instances d = InstancesHelper.getDatasetFromTask(openml, t);            // obtain the dataset
    Pair<Integer, Run> result = RunOpenmlJob.executeTask(openml, new WekaConfig(), taskId, tree);
    Run run = openml.runGet(result.getLeft());
  }    
}
\end{lstlisting}
      \caption{Java, available on Maven Central with artifact id \href{https://mvnrepository.com/artifact/org.openml/openmlweka}{org.openml.openmlweka}}
      \end{subfigure}
    \begin{subfigure}{\textwidth}
\begin{lstlisting}[language=R]
library(OpenML)                                        # requires at least package version 1.8
library(mlr)
lrn = makeLearner('classif.rpart')                     # construct a simple CART classifier
bsuite = getOMLStudy('OpenML-CC18')                    # obtain the benchmark suite
task.ids = extractOMLStudyIds(bsuite, 'task.id')       # obtain the list of suggested tasks
for (task.id in task.ids) {                            # iterate over all tasks 
  task = getOMLTask(task.id)                           # download single OML task
  data = as.data.frame(task)                           # obtain raw data set 
  run = runTaskMlr(task, learner = lrn)                # run constructed learner
  setOMLConfig(apikey = 'FILL_IN_OPENML_API_KEY') 
  upload = uploadOMLRun(run)                           # upload and tag the run
}
\end{lstlisting}
      \caption{R, available on CRAN via package \href{https://CRAN.R-project.org/package=OpenML}{OpenML}} 
    \end{subfigure}
    \caption{Complete code examples, in different programming languages, of how any benchmarking suite (here the `OpenML-CC18' suite) can be downloaded and used to evaluate a given algorithm. The Python code also creates a new benchmark study and shares all  results. Uploading requires a (free) API key.\label{fig:code_examples}
    }
  \end{center}
\end{figure}

\section{How to Use OpenML Benchmarking Suites}

To realize all these benefits, we have developed a series of extensions to the OpenML platform:\footnote{All code is open, BSD-3 licenced, and available on \url{https://github.com/openml}}
\begin{itemize}[noitemsep,leftmargin=*]
    \item We added the concepts of a `benchmark suite' as a collection of \textit{tasks}, and a `benchmark study' as a collection of benchmark results (\textit{runs}) obtained on them.
    \item We added data filtering procedures to the APIs and website that allow researchers to exactly specify the constraints for tasks to be included in a benchmark suite.
    \item We provide scripts and notebooks that facilitate the creation and quality assessment of benchmark suites. For instance, they filter out datasets that are modeled too easily, and hence cannot be used to differentiate between most algorithms (see Section~\ref{createsuites}).
    \item Certain types of datasets, such as multilabel, time series, or artificial datasets, may require additional care. We added collaborative and automated annotation (tagging) to filter such datasets accordingly.
\end{itemize}   

In the following, we discuss the three main use cases for benchmarking suites, i.e., creating new suites, retrieving existing suites, and running benchmarks. We provide code examples on how to retrieve, iterate the contents of a benchmark suite and run machine learning algorithms on it in Figure~\ref{fig:code_examples}.\footnote{More detailed and up-to-date instructions can be found on: \url{https://docs.openml.org/benchmark}}

\subsection{Creating New Suites}
\label{createsuites}
To collect data sets for a new suite, one usually starts by determining a list of constraints that datasets or tasks should adhere to (e.g., have a minimal size, a limited amount of class imbalance, and not be a time series).
This is often an iterative refinement process, during which the distribution of currently selected tasks can be visualized, and any existing benchmarking results on these tasks can be retrieved. An example of this workflow is illustrated in the provided notebook.\footnote{\label{footnote-notebook}Notebooks can be found at \url{https://github.com/openml/benchmark-suites}}
The final selection of tasks can then be used to create a new benchmark suite.
Each benchmark suite is assigned a unique id and an overview webpage with a description and an analysis dashboard (e.g., \url{https://www.openml.org/s/99}).
The description text can be used to describe the goals and design criteria, provide links to external resources, and address any ethical concerns that should be taken into consideration when using the benchmark suite.
We give an exemplary curation protocol in \WithOrWithoutAppendix{Appendix~\ref{sec:protocol}.}{the Appendix.}

\subsection{Retrieving Existing Suites}
Existing benchmark suites can be easily downloaded via any of the OpenML client libraries using its unique id or alias (see Figure~\ref{fig:code_examples}).
The tasks and datasets are all uniformly formatted, and come with extensive meta-data to streamline the execution of benchmarks on them. 
For instance, if a dataset contains missing values, this is indicated in a machine-readable way so that researchers can automatically adjust for this when running their algorithms. 
Datasets can be investigated using exploratory data analysis tools, and existing runs on these tasks can be downloaded and analyzed.

\subsection{Running Benchmarks}
After retrieving the tasks from a suite, new experiments can be conducted locally. As illustrated in Figure~\ref{fig:code_examples}, this is easiest with the readily integrated machine learning libraries, such as scikit-learn~\citep{scikit-learn}, mlr~\citep{Bischl2016} or its successor mlr3~\citep{lang2019mlr3}, and Weka~\citep{Hall2009}. Integrations for deep learning libraries are under development, and we welcome further open source integrations.\footnote{Development is carried out on GitHub. Contributor guidance can be found at \url{https://docs.openml.org}.} Custom code can often be wrapped, e.g., using the scikit-learn interface.

The results of these experiments (runs) can also (optionally) be bundled in a benchmark study and published on OpenML, as illustrated for Python in Figure~\ref{fig:code_examples}. Runs include all experiment details, including hyperparameter configurations, in a structured way. This allows entire communities of scientists to bring together benchmarks of a wide range of algorithms, all evaluated uniformly on the same tasks, in a single place where they can be directly compared on predictive performance and analysed in novel ways. Figure \ref{joyplot} visualizes the results of 3.8 million runs collected on a single benchmarking suite, which we will discuss next.

\section{OpenML-CC18}
\label{cc18}

To demonstrate the functionality of OpenML benchmarking suites, we created a first standard of \numdatasets{} classification tasks built on a carefully curated selection of datasets from the many thousands available on OpenML: the OpenML-CC18. 
It can be used as a drop-in replacement for many typical benchmarking setups. These datasets are deliberately medium-sized for practical reasons. An overview of the benchmark suite can be found at \url{https://www.openml.org/s/99} and \WithOrWithoutAppendix{in Table~\ref{tab:datasets}.}{the Appendix.}
We first describe the design criteria of the OpenML-CC18 before discussing uses of the benchmark and success stories.$^{\ref{footnote-arxiv},}$\footnote{The OpenML-CC18 is the successor of a preliminary benchmarking study called OpenML100, containing 100 classification datasets, and fixes several issues we encountered when working with the OpenML100.}

\begin{figure}[t]
\centering
\includegraphics[width=.9\linewidth]{CC18-joyplot.png}
\caption{Distribution of the scores (average area under ROC curve, weighted by class support) of 3.8 million experiments with thousands of machine learning pipelines, shared on the CC18 benchmark tasks. Some tasks prove harder than others, some have wide score ranges, and for all there exist models that perform poorly (0.5 AUC). Code to reproduce this figure (for any metric) is available on GitHub.$^{\ref{footnote-notebook}}$}
\centering
\label{joyplot}
\end{figure}

\subsection{Design Criteria}
The OpenML-CC18 contains all verified and publicly licenced OpenML datasets until mid-2018 that satisfy a large set of clear requirements for thorough yet practical benchmarking:
\begin{enumerate}[(a),noitemsep, leftmargin=*]
  \item The number of observations is between \num{500} and \num{100000} to focus on medium-sized datasets that can be used to train models on almost any computing hardware.
  \item The dataset has less than 5000 features, counted after one-hot-encoding categorical features (which is the most frequent way to deal with categorical variables), to avoid most memory issues.
  \item The target attribute has at least two classes, with no class of less than 20 observations. This ensures sufficient samples per class per fold when running 10-fold cross-validation experiments.
  \item The ratio of the minority and majority class is above $0.05$ (to eliminate highly imbalanced datasets which require special treatment for both algorithms and evaluation measures).
  \item The dataset is not sparse because not all machine learning models
  can handle them gracefully, this constraint facilitates our goal of wide applicability.
  \item The dataset does not require taking time dependency between samples into account, e.g., time series or data streams, as this is often not implemented in standard machine learning libraries. As a precaution, we also removed datasets where each sample constitutes a single data stream.
  \item The dataset does not require grouped sampling. Such datasets would contain multiple data points for one subject and require that all data points for a subject are put into the same data split for evaluation. We introduce this constraint and the one above to simplify usage of the datasets, as one does not have to use specialized cross-validation procedures.
\end{enumerate}

We also applied several more opinionated criteria to avoid issues with problematic datasets: 
\begin{enumerate}[(a),noitemsep, leftmargin=*]
  \item We strived to remove artificial datasets, as it is hard to reliably assess their difficulty. Admittedly, there is no perfect distinction between artificial and simulated datasets (for example, a lot of phenomena can be simulated that can be as simple as an artificial dataset). Therefore, we removed datasets if we were in doubt of whether they are simulated or artificial.
  \item We removed datasets which are a subset of larger datasets. Allowing subsets would be very subjective, as there is no objective choice of a dataset subset size or a subset of the variables or classes. 
  \item We excluded tasks for which the original target feature has been transformed or changed, e.g., when classes of a categorical target feature were merged or when a continuous target feature (for original regression tasks) was discretized to create a classification task.
  \item We removed datasets without any source or reference. We want to be able to learn more about their intended use and how to interpret learned models, and avoid black box datasets.
\end{enumerate}

Finally, to ensure that datasets are sufficiently challenging, we applied the following restrictions:
\begin{enumerate}[(a),noitemsep, leftmargin=*]
    \item We removed datasets which can be perfectly classified by a single attribute or a decision stump, as they do not allow us to meaningfully compare machine learning algorithms.
  \item We removed datasets where a decision tree could achieve 100\% accuracy on a 10-fold cross-validation task, to remove datasets which can be solved by a simple algorithm which is prone to overfitting training data. We found that this is a good indicator of too easy datasets. Obviously, other datasets will appear easy for several algorithms, and we aim to learn more about the characteristics of such datasets in future studies.
\end{enumerate}

We created the OpenML-CC18 as a first, practical benchmark suite.
In hindsight, we acknowledge that our initial selection still contains several mistakes. Concretely, \emph{sick} is a newer version of the \emph{hypothyroid} dataset with several classes merged, \emph{electricity} has time-related features, \emph{balance\_scale} is an artificial dataset and \emph{mnist\_784} requires grouping samples by writers. We will correct these mistakes in new versions of this suite and also screen the more than 900 new datasets that were uploaded to OpenML since the creation of the OpenML-CC18. Moreover, to avoid the risk of overfitting on a specific benchmark, and to include feedback from the community, we plan to create a dynamic benchmark with regular release updates that evolve with the machine learning field.
We want to clarify that while we include some datasets which may have ethical concerns, we do not expect this to have an impact if the suite is used responsibly (i.e., the benchmark suite is used for its intended purpose of benchmarking algorithms, and not to construct models to be used in real-world applications).

\subsection{Usage of the OpenML-CC18}
\label{sssec:use-of-cc18}

The OpenML-CC18 has been acknowledged and used in various studies.$^{\ref{footnote-arxiv}}$ For instance, \cite{van2020missing} used it to study iterative imputation algorithms for imputing missing values, \cite{KonEtAl20} used it to develop methods to improve upon uncertainty quantification of machine learning classifiers and \cite{debie-arxiv2020a} introduced deep networks for learning meta-features, which they computed for all OpenML-CC18 datasets. In some cases, the authors needed a filtered subset of the OpenML-CC18, which is natively supported in most OpenML clients. Other uses of the OpenML-CC18 include interpreting its multiclass datasets as multi-arm contextual bandit problems~\citep{bibaut-arxiv2021a,bibaut-arxiv2021b} and using the individual columns to test quantile sketch algorithms~\citep{mitchell-acm2021a}.

\cite{cardoso2021data} claim that the machine learning community has a strong focus on algorithmic development, and advocate a more data-centric approach. To this end, they studied the OpenML-CC18 utilizing methods from Item Response Theory to determine which datasets are hard for many classifiers. After analyzing 60 of its datasets (excluding the largest), they find that the OpenML-CC18 consists of both easy and hard datasets. They conclude that the suite is not very challenging as a whole, but that it includes many appropriate datasets to distinguish good classifiers from bad classifiers, and then propose two subsets: one that can be considered challenging, and one subset to replicate the behavior of the full suite. The careful analysis and subsequent proposed updates are a nice example of the natural evolution of benchmarking suites. 

For completeness, we also briefly mention uses of OpenML100, a predecessor of the OpenML-CC18 that includes 100 datasets and less strict constraints. \cite{fabra2020family} use this suite to build a taxonomy of classifiers. They argue that the taxonomies provided by the community can be misleading, and therefore learn taxonomies to cluster classifiers based on predictive behavior. \cite{Rijn2018} and \cite{probst-jmlr2019a} used it to quantify the hyperparameter importance of machine learning algorithms, while \cite{probst2019hyperparameters} used it to learn the best strategy for tuning random forest based on large-scale experiments (although \cite{probst-jmlr2019a} and \cite{probst2019hyperparameters} use only the binary datasets without missing values).

Based upon these works, we conclude that the OpenML-CC18 is being used to facilitate very diverse directions of machine learning research. 

\section{Further OpenML Benchmarking Suites}\label{sec:further_suites}

We now review other OpenML benchmarking suites. For this, we focus on AutoML benchmarking suites, but also provide examples of others.

\subsection{The AutoML Benchmark Suite}
\label{automlbenchmark}
The AutoML benchmark~\citep{gijsbers2019open} also makes use of an OpenML benchmark suite to evaluate AutoML tools in a reproducible manner.
Combined with code to automatically run experiments, any of the integrated AutoML tools can be evaluated on any suitable OpenML task or suite directly from the command line.

\subsubsection{Benchmark Suite Design}
The AutoML benchmark explicitly sources part of their datasets from the OpenML-CC18, but also includes datasets used in AutoML competitions (primarily \cite{guyon-automl2019a}) or previous comparisons of AutoML systems.
A step-by-step list of recreating the benchmark suite does not exist, but general guidelines are provided.
Since the original release in 2019, the AutoML benchmark has been extending their selection of datasets.\footnote{Announcement of the new suites: \url{https://github.com/openml/automlbenchmark/issues/187}}$^,$\footnote{\url{https://www.openml.org/s/{218, 269, 271}} are the original, regression, and expanded suite, respectively}
In the discussion below, aspects which are specific to the newer selection are indicated with an asterisk (*).

The suite shares some of its design criteria with OpenML-CC18, such as the minimum number of instances, as well as the exclusion of artificial datasets and those which require grouped sampling.
However, it loosens some other restrictions specifically because of the assumption that AutoML tools should be able to deal with additional complexities:
\begin{enumerate}[(a),noitemsep, leftmargin=*]
    \item There is no limit to \num{100000} instances or \num{5000} features, tools can restrict themselves to learners which scale well or use, e.g., low-fidelity estimates.
    \item There is no limit for class imbalance, tools can use their preferred techniques to deal with imbalanced data (e.g., SMOTE~\citep{chawla-jair2002a}).
    \item It includes sparse data, though it is currently converted to dense format for tools that don't support sparse data.*
    \item It includes a suite of regression problems.*
\end{enumerate}

Some other restrictions are instead stricter because of the tabular AutoML context:
\begin{enumerate}[(a),noitemsep, leftmargin=*]
    \item The "easy dataset" filter also takes into account results from OpenML across various learners, to try to avoid datasets which need little search beyond algorithm selection.
    \item The number of image classification problems is explicitly restricted, as they are typically better solved with Deep Learning and the benchmark's focus is tabular AutoML tools.
\end{enumerate}

Similar to OpenML-CC18, the AutoML benchmark suite is intended to be regularly updated to reflect modern day challenges and to avoid overfitting.

\subsubsection{Usage of the AutoML Benchmark Suite}
Before the introduction of the AutoML benchmark suite, the closest to an accepted standard for tabular AutoML benchmarking was the set of datasets on which Auto-WEKA was originally evaluated~\citep{thornton2013auto}.
This selection of tasks was still used in, e.g., \cite{wever2018ml} and consisted of 21 problems, a third of which are image classification tasks which are typically not the intended use-case for the AutoML tools.
However, it was by no means a standard. For example, \cite{drori2018alphad3m}, \cite{rakotoarison2018automl} and \cite{gil2018p4ml}, all published at the same workshop, each used different selections of datasets. 

The original AutoML benchmark suite has been used in multiple AutoML publications, either as is \citep{ledell2020h2o, wang2021flaml, feurer2020auto} or with modifications.
Sometimes more datasets are used, as \cite{zoller2021benchmark} combine it with OpenML-CC18 and OpenML100 and \cite{kadra2021regularization} add datasets from UCI and Kaggle.
For the latter, hold-out evaluation is used instead of the suite-defined 10-fold cross-validation.
\cite{erickson2020autogluontabular} use additional datasets from Kaggle competitions to compare directly to solutions proposed by human competitors.

Other times not all datasets in the benchmark suite are used, e.g., \cite{AutoPyTorch} uses all but four big datasets for computational reasons, while \cite{parmentier2019tpot} limit themselves to only four of the big datasets in the suite to assess their method designed for big datasets.
\cite{mohr2021replacing} omitted some datasets because of technical issues.

\subsection{Further Existing OpenML Benchmarking Suites}

OpenML contains other benchmark suites as well, such as the \href{https://www.openml.org/s/225}{OpenML100-friendly} that only contains the subset of the OpenML100 without missing values and with only numerical features, or \href{https://www.openml.org/s/219}{Foreign Exchange} data for machine learning research~\citep{Schut2019}.

We invite the community to create additional benchmarks suites for other tasks besides classification, for larger datasets or more high-dimensional ones, for imbalanced or extremely noisy datasets, as well as for text, time series, and many other types of data. We are confident that benchmarking suites will help standardize evaluation and track progress in many subfields of machine learning, and also intend to create new suites and make it ever easier for others to do so.

\section{Limitations and Future Work}

As benchmarking suites are increasingly being picked up by the machine learning community, we also observed several limitations that should be tackled in future work.

\textbf{Overfitting.} While it has not yet been demonstrated, we assume that as more methods are being evaluated on benchmarking suites, overfitting on fixed suites is increasingly likely. We therefore aim to periodically update existing suites with new datasets that follow the specifications laid out by the benchmark designers (e.g., as done for computer vision research~\citep{recht-icml2019a}) and invite the community to extend existing suites with harder tasks, as done in NLP research~\citep{kiela2021dynabench}.

\textbf{Credit Assignment.} Curating a benchmark is a lot of work, and we have manually inspected and corrected datasets for the OpenML-CC18 over the course of multiple months. It is therefore important to give proper credit to everyone involved in creating benchmarking suites, for example by somehow making benchmarking suites citable.

\textbf{Automating the curation of useful suites.} We are not aware of any related work that describes how to curate machine learning benchmark suites. In this paper we have defined benchmarking suites by formalizing objective, but also more subjective constraints. Providing automated ways to create high quality, diverse and realistic benchmarking suites is thus an important, open research question. Additionally, post-hoc research, such as the one conducted by \cite{cardoso2021data}, is important to check the validity of benchmarking suites, and we hope for more such techniques to be developed and also to become applicable during the suite design process. 

\textbf{Computational issues.} While studying applications of the OpenML-CC18 in Section~\ref{sssec:use-of-cc18} we realized that even though we consciously focused on mid-size datasets, some larger ones still incurred too high computational load, so some researchers have used subsets of the OpenML-CC18 in their work. Future suites could more carefully trade off the completeness of benchmarking suites and computational issues, for example by choosing representative subsets~\citep{cardoso2021data}.

\textbf{Breadth of current benchmarking suites.} On the other hand, many researchers are interested in benchmarking larger (deep learning) models on larger datasets from many domains (including language and vision). We are working on ways to enable the creation of such benchmarking suites as well, and welcome further involvement from the community.

\textbf{Specification of resource constraints.} The task and suite specifications do not yet allow for constraints on resources, e.g., memory or time limits. Specific benchmark studies could impose identical hardware requirements, e.g., to compare running times. Where requiring identical hardware is impractical, general constraints would ensure results are more comparable when multiple people run their experiments on a suite. Explicit constraints also help interpret earlier results.

\textbf{Disclosure of ethical issues} We currently encourage creators to disclose any ethical concerns with datasets in their benchmark suite in its description. In the future we want to support this natively on a dataset level (e.g., by integrating datasheets~\citep{gebru2018datasheets}) and benchmark suite level (by providing a dedicated information field).

\section{Conclusion}
Our goal is to simplify the creation of well-designed benchmarks to push machine learning research forward. 
More than just creating and sharing benchmarks, we want to allow anyone to effortlessly run and publish their own benchmarking results and organize them online in a single place where they can be easily explored, downloaded, shared, compared, and analyzed. 
We created a new benchmarking layer on the OpenML platform that allows scientists to do all the above with just a few lines of code. We then introduced the OpenML-CC18, a benchmark suite created with these tools for general classification benchmarking. 

The use of suites is further motivated by a closer look at the AutoML benchmark suite. We also reviewed how other scientists have adopted these benchmarking suites in their own work, from which it becomes clear that a continuous conversation with the research community is essential to evolve benchmarks and make them better and more useful over time. We hope that this work will unleash a rapid evolution of benchmarks suites and large-scale studies that teach us more about machine learning than any single study could.

\noindent \textbf{Acknowledgements} This work has partly been funded by the German Federal Ministry of Education and Research (BMBF) under grant no. 01IS18036A, by the Deutsche Forschungsgemeinschaft (DFG, German Research Foundation) – 460135501 (NFDI project MaRDI), by the European Research Council (ERC) under the European Union's Horizon 2020 research and innovation programme under grant no.\ 716721 (Beyond BlackBox) and 952215 (TAILOR), through grant \#2015/03986-0 from the S{\~a}o Paulo Research Foundation (FAPESP), by AFRL and DARPA under contract FA8750-17-C-0141, as well as through the Priority Programme Autonomous Learning (SPP 1527, grant HU~1900/3-1) and Collaborative Research Center SFB~876/A3 from the German Research Foundation (DFG). In addition, we would like to thank Andreas Müller for his feedback on the OpenML100.

\bibliography{openml100.bib}

\begin{thebibliography}{56}
\providecommand{\natexlab}[1]{#1}
\providecommand{\url}[1]{\texttt{#1}}
\expandafter\ifx\csname urlstyle\endcsname\relax
  \providecommand{\doi}[1]{doi: #1}\else
  \providecommand{\doi}{doi: \begingroup \urlstyle{rm}\Url}\fi

\bibitem[Aha(1992)]{Aha:1992p455}
D.~W. Aha.
\newblock Generalizing from case studies: A case study.
\newblock \emph{Proceedings of the International Conference on Machine Learning
  (ICML)}, pages 1--10, 1992.

\bibitem[Alcala et~al.(2010)Alcala, Fernandez, Luengo, Derrac, Garcia, Sanchez,
  and Herrera]{KEEL}
J.~Alcala, A.~Fernandez, J.~Luengo, J.~Derrac, S.~Garcia, L.~Sanchez, and
  F.~Herrera.
\newblock Keel datamining software tool: Data set repository, integration of
  algorithms and experimental analysis framework.
\newblock \emph{Journal of Multiple-Valued Logic and Soft Computing},
  17\penalty0 (2-3):\penalty0 255--287, 2010.

\bibitem[Bergstra et~al.(2015)Bergstra, Pinto, and Cox]{bergstra2015skdata}
J.~Bergstra, N.~Pinto, and D.~Cox.
\newblock Skdata: data sets and algorithm evaluation protocols in python.
\newblock \emph{Computational Science \& Discovery}, 8\penalty0 (1), 2015.

\bibitem[Bibaut et~al.(2021{\natexlab{a}})Bibaut, Chambaz, Dimakopoulou,
  Kallus, and van~der Laan]{bibaut-arxiv2021a}
A.~Bibaut, A.~Chambaz, M.~Dimakopoulou, N.~Kallus, and M.~van~der Laan.
\newblock Risk minimization from adaptively collected data: Guarantees for
  supervised and policy learning.
\newblock \emph{arXiv:2106.01723 [stat.ML]}, 2021{\natexlab{a}}.

\bibitem[Bibaut et~al.(2021{\natexlab{b}})Bibaut, Chambaz, Dimakopoulou,
  Kallus, and van~der Laan]{bibaut-arxiv2021b}
A.~Bibaut, A.~Chambaz, M.~Dimakopoulou, N.~Kallus, and M.~van~der Laan.
\newblock Post-contextual-bandit inference.
\newblock \emph{arXiv:2106.00418 [stat.ML]}, 2021{\natexlab{b}}.

\bibitem[Bischl et~al.(2016{\natexlab{a}})Bischl, Kerschke, Kotthoff, Lindauer,
  Malitsky, Frech\'{e}tte, Hoos, Hutter, Leyton-Brown, Tierney, and
  Vanschoren]{bischl-aij16a}
B.~Bischl, P.~Kerschke, L.~Kotthoff, M.~Lindauer, Y.~Malitsky,
  A.~Frech\'{e}tte, H.~Hoos, F.~Hutter, K.~Leyton-Brown, K.~Tierney, and
  J.~Vanschoren.
\newblock {ASlib}: A benchmark library for algorithm selection.
\newblock \emph{Artificial Intelligence}, 237:\penalty0 41--58,
  2016{\natexlab{a}}.

\bibitem[Bischl et~al.(2016{\natexlab{b}})Bischl, Lang, Kotthoff, Schiffner,
  Richter, Studerus, Casalicchio, and Jones]{Bischl2016}
B.~Bischl, M.~Lang, L.~Kotthoff, J.~Schiffner, J.~Richter, E.~Studerus,
  G.~Casalicchio, and Z.~M. Jones.
\newblock mlr: Machine learning in {R}.
\newblock \emph{Journal of Machine Learning Research}, 17\penalty0 (170),
  2016{\natexlab{b}}.

\bibitem[Brockman et~al.(2016)Brockman, Cheung, Pettersson, Schneider,
  Schulman, Tang, and Zaremba]{openai_gym}
G.~Brockman, V.~Cheung, L.~Pettersson, J.~Schneider, J.~Schulman, J.~Tang, and
  W.~Zaremba.
\newblock {OpenAI} {Gym}.
\newblock \emph{arXiv:1606.01540 [cs.LG]}, 2016.

\bibitem[Cardoso et~al.(2021)Cardoso, Santos, Franc{\^e}s, Prud{\^e}ncio, and
  Alves]{cardoso2021data}
L.~F. Cardoso, V.~C. Santos, R.~S.~K. Franc{\^e}s, R.~B. Prud{\^e}ncio, and
  R.~C. Alves.
\newblock Data vs classifiers, who wins?
\newblock \emph{arXiv:2107.07451 [cs.LG]}, 2021.

\bibitem[Casalicchio et~al.(2017)Casalicchio, Bossek, Lang, Kirchhoff,
  Kerschke, Hofner, Seibold, Vanschoren, and Bischl]{Casalicchio2017}
G.~Casalicchio, J.~Bossek, M.~Lang, D.~Kirchhoff, P.~Kerschke, B.~Hofner,
  H.~Seibold, J.~Vanschoren, and B.~Bischl.
\newblock {OpenML}: An {R} package to connect to the machine learning platform
  {OpenML}.
\newblock \emph{Computational Statistics}, 34\penalty0 (3):\penalty0 977--991,
  2017.

\bibitem[Chang and Lin(2011)]{chang2011libsvm}
C.~C. Chang and C.~J. Lin.
\newblock {LIBSVM}: A library for support vector machines.
\newblock \emph{ACM Transactions on Intelligent Systems and Technology (TIST)},
  2\penalty0 (3):\penalty0 27, 2011.

\bibitem[Chawla et~al.(2002)Chawla, Bowyer, Hall, and
  Kegelmeyer]{chawla-jair2002a}
N.~V. Chawla, K.~W. Bowyer, L.~O. Hall, and W.~P. Kegelmeyer.
\newblock Smote: Synthetic minority over-sampling technique.
\newblock \emph{Journal of Artificial Intelligence Research}, 16\penalty0
  (1):\penalty0 321–357, 2002.

\bibitem[Chen et~al.(2015)Chen, Keogh, Hu, Begum, Bagnall, Mueen, and
  Batista]{UCRArchive}
Y.~Chen, E.~Keogh, B.~Hu, N.~Begum, A.~Bagnall, A.~Mueen, and G.~Batista.
\newblock The {UCR} time series classification archive, July 2015.
\newblock \url{www.cs.ucr.edu/~eamonn/time_series_data/}.

\bibitem[De~Bie et~al.(2020)De~Bie, Rakotoarison, Peyré, and
  Sebag]{debie-arxiv2020a}
G.~De~Bie, H.~Rakotoarison, G.~Peyré, and M.~Sebag.
\newblock Distribution-based invariant deep networks for learning
  meta-features.
\newblock \emph{arXiv:2006.13708 [stat.ML]}, 2020.

\bibitem[Dheeru and Taniskidou(2017)]{UCI:2017}
D.~Dheeru and E.~K. Taniskidou.
\newblock {UCI} machine learning repository, 2017.
\newblock URL \url{http://archive.ics.uci.edu/ml}.

\bibitem[Drori et~al.(2018)Drori, Krishnamurthy, Rampin, Louren{\c{c}}o, One,
  Cho, Silva, and Freire]{drori2018alphad3m}
I.~Drori, Y.~Krishnamurthy, R.~Rampin, R.~Louren{\c{c}}o, J.~One, K.~Cho,
  C.~Silva, and J.~Freire.
\newblock Alphad3m: Machine learning pipeline synthesis.
\newblock In \emph{5th ICML Workshop on Automated Machine Learning (AutoML)},
  2018.

\bibitem[Erickson et~al.(2020)Erickson, Mueller, Shirkov, Zhang, Larroy, Li,
  and Smola]{erickson2020autogluontabular}
N.~Erickson, J.~Mueller, A.~Shirkov, H.~Zhang, P.~Larroy, M.~Li, and A.~Smola.
\newblock Autogluon-tabular: Robust and accurate automl for structured data.
\newblock \emph{arXiv:2003.06505 [stat.ML]}, 2020.

\bibitem[Fabra-Boluda et~al.(2020)Fabra-Boluda, Ferri, Mart{\'\i}nez-Plumed,
  Hern{\'a}ndez-Orallo, and Ram{\'\i}rez-Quintana]{fabra2020family}
R.~Fabra-Boluda, C.~Ferri, F.~Mart{\'\i}nez-Plumed, J.~Hern{\'a}ndez-Orallo,
  and M.~J. Ram{\'\i}rez-Quintana.
\newblock Family and prejudice: A behavioural taxonomy of machine learning
  techniques.
\newblock In \emph{ECAI 2020 - 24th European Conference on Artificial
  Intelligence}, pages 1135--1142. IOS Press, 2020.

\bibitem[Feurer et~al.(2021{\natexlab{a}})Feurer, Eggensperger, Falkner,
  Lindauer, and Hutter]{feurer2020auto}
M.~Feurer, K.~Eggensperger, S.~Falkner, M.~Lindauer, and F.~Hutter.
\newblock Auto-sklearn 2.0: Hands-free automl via meta-learning.
\newblock \emph{arXiv:2007.04074 [cs.LG]}, 2021{\natexlab{a}}.

\bibitem[Feurer et~al.(2021{\natexlab{b}})Feurer, van Rijn, Kadra, Gijsbers,
  Mallik, Ravi, Müller, Vanschoren, and Hutter]{feurer-jmlr2021a}
M.~Feurer, J.~N. van Rijn, A.~Kadra, P.~Gijsbers, N.~Mallik, S.~Ravi,
  A.~Müller, J.~Vanschoren, and F.~Hutter.
\newblock Openml-python: an extensible python api for openml.
\newblock \emph{Journal of Machine Learning Research}, 22\penalty0
  (100):\penalty0 1--5, 2021{\natexlab{b}}.

\bibitem[Gebru et~al.(2018)Gebru, Morgenstern, Vecchione, Vaughan, Wallach,
  Daum{\'e}~III, and Crawford]{gebru2018datasheets}
T.~Gebru, J.~Morgenstern, B.~Vecchione, J.~W. Vaughan, H.~Wallach,
  H.~Daum{\'e}~III, and K.~Crawford.
\newblock Datasheets for datasets.
\newblock \emph{arXiv:1803.09010 [cs.DB]}, 2018.

\bibitem[Gijsbers et~al.(2019)Gijsbers, LeDell, Thomas, Poirier, Bischl, and
  Vanschoren]{gijsbers2019open}
P.~Gijsbers, E.~LeDell, J.~Thomas, S.~Poirier, B.~Bischl, and J.~Vanschoren.
\newblock An open source automl benchmark.
\newblock In \emph{6th ICML Workshop on Automated Machine Learning (AutoML)},
  2019.

\bibitem[Gil et~al.(2018)Gil, Yao, Ratnakar, Garijo, Ver~Steeg, Szekely,
  Brekelmans, Kejriwal, Luo, and Huang]{gil2018p4ml}
Y.~Gil, K.-T. Yao, V.~Ratnakar, D.~Garijo, G.~Ver~Steeg, P.~Szekely,
  R.~Brekelmans, M.~Kejriwal, F.~Luo, and I.-H. Huang.
\newblock P4ml: A phased performance-based pipeline planner for automated
  machine learning.
\newblock In \emph{5th ICML Workshop on Automated Machine Learning (AutoML)},
  2018.

\bibitem[Guyon et~al.(2019)Guyon, Sun-Hosoya, Boull{\'e}, Escalante, Escalera,
  Liu, Jajetic, Ray, Saeed, Sebag, Statnikov, Tu, and
  Viegas]{guyon-automl2019a}
I.~Guyon, L.~Sun-Hosoya, M.~Boull{\'e}, H.~J. Escalante, S.~Escalera, Z.~Liu,
  D.~Jajetic, B.~Ray, M.~Saeed, M.~Sebag, A.~Statnikov, W.-W. Tu, and
  E.~Viegas.
\newblock Analysis of the automl challenge series 2015--2018.
\newblock In F.~Hutter, L.~Kotthoff, and J.~Vanschoren, editors,
  \emph{Automated Machine Learning: Methods, Systems, Challenges}, pages
  177--219. Springer International Publishing, 2019.

\bibitem[Hall et~al.(2009)Hall, Frank, Holmes, Pfahringer, Reutemann, and
  Witten]{Hall2009}
M.~Hall, E.~Frank, G.~Holmes, B.~Pfahringer, P.~Reutemann, and I.~H. Witten.
\newblock The {WEKA} data mining software: An update.
\newblock \emph{ACM SIGKDD Explorations Newsletter}, 11\penalty0 (1):\penalty0
  10--18, 2009.

\bibitem[Hansen et~al.(2020)Hansen, Auger, Ros, Mersman, Tu{\v s}ar, and
  Brockhoff]{hansen-oms20}
N.~Hansen, A.~Auger, R.~Ros, O.~Mersman, T.~Tu{\v s}ar, and D.~Brockhoff.
\newblock {COCO}: A platform for comparing continuous optimizers in a black-box
  setting.
\newblock \emph{Optimization Methods and Software}, 2020.

\bibitem[Hutson(2018)]{Hutson2018}
M.~Hutson.
\newblock Missing data hinder replication of artificial intelligence studies.
\newblock \emph{Science News}, 2018.
\newblock URL
  \url{https://www.science.org/content/article/missing-data-hinder-replication-artificial-intelligence-studies}.

\bibitem[Kadra et~al.(2021)Kadra, Lindauer, Hutter, and
  Grabocka]{kadra2021regularization}
A.~Kadra, M.~Lindauer, F.~Hutter, and J.~Grabocka.
\newblock Regularization is all you need: Simple neural nets can excel on
  tabular data.
\newblock \emph{arXiv:2106.11189 [cs.LG]}, 2021.

\bibitem[Keogh and Kasetty(2003)]{Keogh:2003p4930}
E.~Keogh and S.~Kasetty.
\newblock On the need for time series data mining benchmarks: A survey and
  empirical demonstration.
\newblock \emph{Data Mining and Knowledge Discovery}, 7\penalty0 (4):\penalty0
  349--371, 2003.

\bibitem[Kiela et~al.(2021)Kiela, Bartolo, Nie, Kaushik, Geiger, Wu, Vidgen,
  Prasad, Singh, Ringshia, Ma, Thrush, Riedel, Waseem, Stenetorp, Jia, Bansal,
  Potts, and Williams]{kiela2021dynabench}
D.~Kiela, M.~Bartolo, Y.~Nie, D.~Kaushik, A.~Geiger, Z.~Wu, B.~Vidgen,
  G.~Prasad, A.~Singh, P.~Ringshia, Z.~Ma, T.~Thrush, S.~Riedel, Z.~Waseem,
  P.~Stenetorp, R.~Jia, M.~Bansal, C.~Potts, and A.~Williams.
\newblock Dynabench: Rethinking benchmarking in {NLP}.
\newblock In \emph{Proceedings of the 2021 Conference of the North American
  Chapter of the Association for Computational Linguistics: Human Language
  Technologies}, pages 4110--4124. Association for Computational Linguistics,
  2021.

\bibitem[K{\"o}nig et~al.(2020)K{\"o}nig, Hoos, and van Rijn]{KonEtAl20}
M.~K{\"o}nig, H.~H. Hoos, and J.~N. van Rijn.
\newblock Towards algorithm-agnostic uncertainty estimation: Predicting
  classification error in an automated machine learning setting.
\newblock In \emph{7th ICML Workshop on Automated Machine Learning (AutoML)},
  2020.

\bibitem[Lang et~al.(2019)Lang, Binder, Richter, Schratz, Pfisterer, Coors, Au,
  Casalicchio, Kotthoff, and Bischl]{lang2019mlr3}
M.~Lang, M.~Binder, J.~Richter, P.~Schratz, F.~Pfisterer, S.~Coors, Q.~Au,
  G.~Casalicchio, L.~Kotthoff, and B.~Bischl.
\newblock mlr3: A modern object-oriented machine learning framework in r.
\newblock \emph{Journal of Open Source Software}, 4\penalty0 (44):\penalty0
  1903, 2019.

\bibitem[LeDell and Poirier(2020)]{ledell2020h2o}
E.~LeDell and S.~Poirier.
\newblock H2o automl: Scalable automatic machine learning.
\newblock In \emph{7th ICML Workshop on Automated Machine Learning (AutoML)},
  2020.

\bibitem[Mitchell et~al.(2021)Mitchell, Frank, and Holmes]{mitchell-acm2021a}
R.~Mitchell, E.~Frank, and G.~Holmes.
\newblock An empirical study of moment estimators for quantile approximation.
\newblock \emph{ACM Transactions on Database Systems}, 46\penalty0 (1), 2021.

\bibitem[Mohr and Wever(2021)]{mohr2021replacing}
F.~Mohr and M.~Wever.
\newblock Replacing the ex-def baseline in automl by naive automl.
\newblock In \emph{8th ICML Workshop on Automated Machine Learning (AutoML)},
  2021.

\bibitem[Mohr et~al.(2018)Mohr, Wever, and H{\"{u}}llermeier]{wever2018ml}
F.~Mohr, M.~Wever, and E.~H{\"{u}}llermeier.
\newblock Ml-plan: Automated machine learning via hierarchical planning.
\newblock \emph{Machine Learning}, 107\penalty0 (8-10):\penalty0 1495--1515,
  2018.

\bibitem[Narayan et~al.(2021)Narayan, Molino, Goel, Neiswanger, and
  Re]{narayan-neurips2021a}
A.~Narayan, P.~Molino, K.~Goel, W.~Neiswanger, and C.~Re.
\newblock Personalized benchmarking with the ludwig benchmarking toolkit.
\newblock In \emph{Proceedings of the Neural Information Processing Systems
  Track on Datasets and Benchmarks}, 2021.

\bibitem[Olson et~al.(2017)Olson, {La Cava}, Orzechowski, Urbanowicz, and
  Moore]{pmlb}
R.~S. Olson, W.~{La Cava}, P.~Orzechowski, R.~J. Urbanowicz, and J.~H. Moore.
\newblock {PMLB}: A large benchmark suite for machine learning evaluation and
  comparison.
\newblock \emph{BioData Mining}, 10\penalty0 (36), 2017.

\bibitem[Parmentier et~al.(2019)Parmentier, Nicol, Jourdan, and
  Kessaci]{parmentier2019tpot}
L.~Parmentier, O.~Nicol, L.~Jourdan, and M.-E. Kessaci.
\newblock {TPOT-SH}: A faster optimization algorithm to solve the automl
  problem on large datasets.
\newblock In \emph{2019 IEEE 31st International Conference on Tools with
  Artificial Intelligence (ICTAI)}, pages 471--478. IEEE, 2019.

\bibitem[Pedersen(2008)]{Pedersen:2008p12980}
T.~Pedersen.
\newblock Empiricism is not a matter of faith.
\newblock \emph{Computational Linguistics}, 34:\penalty0 465--470, 2008.

\bibitem[Pedregosa et~al.(2011)Pedregosa, Varoquaux, Gramfort, Michel, Thirion,
  Grisel, Blondel, Prettenhofer, Weiss, Dubourg, Vanderplas, Passos,
  Cournapeau, Brucher, Perrot, and Duchesnay]{scikit-learn}
F.~Pedregosa, G.~Varoquaux, A.~Gramfort, V.~Michel, B.~Thirion, O.~Grisel,
  M.~Blondel, P.~Prettenhofer, R.~Weiss, V.~Dubourg, J.~Vanderplas, A.~Passos,
  D.~Cournapeau, M.~Brucher, M.~Perrot, and E.~Duchesnay.
\newblock Scikit-learn: Machine learning in {P}ython.
\newblock \emph{Journal of Machine Learning Research}, 12:\penalty0 2825--2830,
  2011.

\bibitem[Probst et~al.(2019{\natexlab{a}})Probst, Boulesteix, and
  Bischl]{probst-jmlr2019a}
P.~Probst, A.-L. Boulesteix, and B.~Bischl.
\newblock Tunability: Importance of hyperparameters of machine learning
  algorithms.
\newblock \emph{Journal of Machine Learning Research}, 20\penalty0
  (53):\penalty0 1--32, 2019{\natexlab{a}}.

\bibitem[Probst et~al.(2019{\natexlab{b}})Probst, Wright, and
  Boulesteix]{probst2019hyperparameters}
P.~Probst, M.~N. Wright, and A.-L. Boulesteix.
\newblock Hyperparameters and tuning strategies for random forest.
\newblock \emph{Wiley Interdisciplinary Reviews: Data Mining and Knowledge
  Discovery}, 9\penalty0 (3):\penalty0 e1301, 2019{\natexlab{b}}.

\bibitem[Rakotoarison et~al.(2019)Rakotoarison, Schoenauer, and
  Sebag]{rakotoarison2018automl}
H.~Rakotoarison, M.~Schoenauer, and M.~Sebag.
\newblock Automated machine learning with monte-carlo tree search.
\newblock In \emph{Proceedings of the Twenty-Eighth International Joint
  Conference on Artificial Intelligence}, pages 3296--3303, 2019.

\bibitem[Recht et~al.(2019)Recht, Roelofs, Schmidt, and
  Shankar]{recht-icml2019a}
B.~Recht, R.~Roelofs, L.~Schmidt, and V.~Shankar.
\newblock Do {I}mage{N}et classifiers generalize to {I}mage{N}et?
\newblock In K.~Chaudhuri and R.~Salakhutdinov, editors, \emph{Proceedings of
  the 36th International Conference on Machine Learning}, volume~97, pages
  5389--5400, 2019.

\bibitem[Schut et~al.(2019)Schut, van Rijn, and Hoos]{Schut2019}
F.~Schut, J.~N. van Rijn, and H.~Hoos.
\newblock Towards automated technical analysis for foreign exchange data.
\newblock In \emph{Workshop on Automating Data Science @ ECML/PKDD}, 2019.

\bibitem[Sculley et~al.(2018)Sculley, Snoek, Wiltschko, and
  Rahimi]{sculley2018winner}
D.~Sculley, J.~Snoek, A.~Wiltschko, and A.~Rahimi.
\newblock Winner's curse? on pace, progress, and empirical rigor.
\newblock In \emph{Workshop of the International Conference on Representation
  Learning (ICLR)}, 2018.

\bibitem[Thornton et~al.(2013)Thornton, Hutter, Hoos, and
  Leyton-Brown]{thornton2013auto}
C.~Thornton, F.~Hutter, H.~H. Hoos, and K.~Leyton-Brown.
\newblock Auto-weka: Combined selection and hyperparameter optimization of
  classification algorithms.
\newblock In \emph{Proceedings of the 19th ACM SIGKDD international conference
  on Knowledge discovery and data mining}, pages 847--855, 2013.

\bibitem[Tsoumakas et~al.(2011)Tsoumakas, {Spyromitros-Xioufis}, Vilcek, and
  Vlahavas]{mulan}
G.~Tsoumakas, E.~{Spyromitros-Xioufis}, J.~Vilcek, and I.~Vlahavas.
\newblock Mulan: A java library for multi-label learning.
\newblock \emph{Journal of Machine Learning Research}, pages 2411--2414, Jul
  2011.

\bibitem[van Rijn(2016)]{Rijn2016}
J.~N. van Rijn.
\newblock \emph{Massively collaborative machine learning}.
\newblock PhD thesis, Leiden University, 2016.

\bibitem[van Rijn and Hutter(2018)]{Rijn2018}
J.~N. van Rijn and F.~Hutter.
\newblock Hyperparameter importance across datasets.
\newblock In \emph{Proceedings of the 24th ACM SIGKDD International Conference
  on Knowledge Discovery and Data Mining}, pages 2367--2376. ACM, 2018.

\bibitem[Van~Wolputte and Blockeel(2020)]{van2020missing}
E.~Van~Wolputte and H.~Blockeel.
\newblock Missing value imputation with mercs: A faster alternative to
  missforest.
\newblock In \emph{Discovery Science - 23rd International Conference}, volume
  12323 of \emph{Lecture Notes in Computer Science}, pages 502--516. Springer,
  2020.

\bibitem[Vanschoren et~al.(2013)Vanschoren, van Rijn, Bischl, and
  Torgo]{OpenML2013}
J.~Vanschoren, J.~N. van Rijn, B.~Bischl, and L.~Torgo.
\newblock {OpenML}: Networked science in machine learning.
\newblock \emph{SIGKDD Explorations}, 15\penalty0 (2):\penalty0 49--60, 2013.

\bibitem[Wang et~al.(2021)Wang, Wu, Weimer, and Zhu]{wang2021flaml}
C.~Wang, Q.~Wu, M.~Weimer, and E.~Zhu.
\newblock {FLAML}: A fast and lightweight automl library.
\newblock \emph{Proceedings of Machine Learning and Systems}, 3, 2021.

\bibitem[Zimmer et~al.(2021)Zimmer, Lindauer, and Hutter]{AutoPyTorch}
L.~Zimmer, M.~Lindauer, and F.~Hutter.
\newblock Auto-pytorch: Multi-fidelity metalearning for efficient and robust
  autodl.
\newblock \emph{IEEE Transactions on Pattern Analysis and Machine
  Intelligence}, 43\penalty0 (9):\penalty0 3079--3090, 2021.

\bibitem[Z{\"o}ller and Huber(2021)]{zoller2021benchmark}
M.-A. Z{\"o}ller and M.~F. Huber.
\newblock Benchmark and survey of automated machine learning frameworks.
\newblock \emph{Journal of Artificial Intelligence Research}, 70:\penalty0
  409--472, 2021.

\end{thebibliography}
\bibliographystyle{abbrvnat}

\ifPrintAppendix

\newpage
\appendix

\section{OpenML-CC18 dataset list}

\begin{table}[tbph]
\setlength{\tabcolsep}{3pt}
\caption{Datasets included in the OpenML-CC18 benchmarking suite. For each dataset, we show: the OpenML task id, dataset id and name, the number of classes (cl), features (p) and observations (n), as well as the ratio of the minority and majority class sizes (MinMaj).}
\label{tab:datasets}
\centering

\begin{adjustbox}{center, width=8cm, totalheight=10cm}
\begin{tabular}{rrlrrrrr}
\toprule
\textbf{ Data id } & \textbf{Task id} & \textbf{ Name } & \textbf{ cl } & \textbf{ p } & \textbf{ n } & \textbf{MinMaj}\\
\midrule

\midrule
3 & 3 & kr-vs-kp & 2 & 37 & 3196 & 0.91\\
6 & 6 & letter & 26 & 17 & 20000 & 0.90\\
11 & 11 & balance-scale & 3 & 5 & 625 & 0.17\\
12 & 12 & mfeat-factors & 10 & 217 & 2000 & 1.00\\
14 & 14 & mfeat-fourier & 10 & 77 & 2000 & 1.00\\
\addlinespace
15 & 15 & breast-w & 2 & 10 & 699 & 0.53\\
16 & 16 & mfeat-karhunen & 10 & 65 & 2000 & 1.00\\
18 & 18 & mfeat-morphological & 10 & 7 & 2000 & 1.00\\
22 & 22 & mfeat-zernike & 10 & 48 & 2000 & 1.00\\
23 & 23 & cmc & 3 & 10 & 1473 & 0.53\\
\addlinespace
28 & 28 & optdigits & 10 & 65 & 5620 & 0.97\\
29 & 29 & credit-approval & 2 & 16 & 690 & 0.80\\
31 & 31 & credit-g & 2 & 21 & 1000 & 0.43\\
32 & 32 & pendigits & 10 & 17 & 10992 & 0.92\\
37 & 37 & diabetes & 2 & 9 & 768 & 0.54\\
\addlinespace
38 & 3021 & sick & 2 & 30 & 3772 & 0.07\\
44 & 43 & spambase & 2 & 58 & 4601 & 0.65\\
46 & 45 & splice & 3 & 62 & 3190 & 0.46\\
50 & 49 & tic-tac-toe & 2 & 10 & 958 & 0.53\\
54 & 53 & vehicle & 4 & 19 & 846 & 0.91\\
\addlinespace
151 & 219 & electricity & 2 & 9 & 45312 & 0.74\\
182 & 2074 & satimage & 6 & 37 & 6430 & 0.41\\
188 & 2079 & eucalyptus & 5 & 20 & 736 & 0.49\\
300 & 3481 & isolet & 26 & 618 & 7797 & 0.99\\
307 & 3022 & vowel & 11 & 13 & 990 & 1.00\\
\addlinespace
458 & 3549 & analcatdata\_authorship & 4 & 71 & 841 & 0.17\\
469 & 3560 & analcatdata\_dmft & 6 & 5 & 797 & 0.79\\
554 & 3573 & mnist\_784 & 10 & 785 & 70000 & 0.80\\
1049 & 3902 & pc4 & 2 & 38 & 1458 & 0.14\\
1050 & 3903 & pc3 & 2 & 38 & 1563 & 0.11\\
\addlinespace
1053 & 3904 & jm1 & 2 & 22 & 10885 & 0.24\\
1063 & 3913 & kc2 & 2 & 22 & 522 & 0.26\\
1067 & 3917 & kc1 & 2 & 22 & 2109 & 0.18\\
1068 & 3918 & pc1 & 2 & 22 & 1109 & 0.07\\
1461 & 14965 & bank-marketing & 2 & 17 & 45211 & 0.13\\
\addlinespace
1462 & 10093 & banknote-authentication & 2 & 5 & 1372 & 0.80\\

\bottomrule
\end{tabular}
~~~~~
\begin{tabular}{rrlrrrrr}
\toprule
\textbf{ Data id } & \textbf{ Task id } & \textbf{ Name } &  \textbf{ cl } & \textbf{ p } & \textbf{ n } & \textbf{MinMaj}\\
\midrule 

\midrule
1464 & 10101 & blood-transfusion-service-center & 2 & 5 & 748 & 0.31\\
1468 & 9981 & cnae-9 & 9 & 857 & 1080 & 1.00\\
1475 & 9985 & first-order-theorem-proving & 6 & 52 & 6118 & 0.19\\
1478 & 14970 & har & 6 & 562 & 10299 & 0.72\\
1480 & 9971 & ilpd & 2 & 11 & 583 & 0.40\\
\addlinespace
1485 & 9976 & madelon & 2 & 501 & 2600 & 1.00\\
1486 & 9977 & nomao & 2 & 119 & 34465 & 0.40\\
1487 & 9978 & ozone-level-8hr & 2 & 73 & 2534 & 0.07\\
1489 & 9952 & phoneme & 2 & 6 & 5404 & 0.42\\
1494 & 9957 & qsar-biodeg & 2 & 42 & 1055 & 0.51\\
\addlinespace
1497 & 9960 & wall-robot-navigation & 4 & 25 & 5456 & 0.15\\
1501 & 9964 & semeion & 10 & 257 & 1593 & 0.96\\
1510 & 9946 & wdbc & 2 & 31 & 569 & 0.59\\
1590 & 7592 & adult & 2 & 15 & 48842 & 0.31\\
4134 & 9910 & Bioresponse & 2 & 1777 & 3751 & 0.84\\
\addlinespace
4534 & 14952 & PhishingWebsites & 2 & 31 & 11055 & 0.80\\
4538 & 14969 & GesturePhaseSegmentationProcessed & 5 & 33 & 9873 & 0.34\\
6332 & 14954 & cylinder-bands & 2 & 40 & 540 & 0.73\\
23381 & 125920 & dresses-sales & 2 & 13 & 500 & 0.72\\
23517 & 167120 & numerai28.6 & 2 & 22 & 96320 & 0.98\\
\addlinespace
40499 & 125922 & texture & 11 & 41 & 5500 & 1.00\\
40668 & 146195 & connect-4 & 3 & 43 & 67557 & 0.15\\
40670 & 167140 & dna & 3 & 181 & 3186 & 0.46\\
40701 & 167141 & churn & 2 & 21 & 5000 & 0.16\\
40923 & 167121 & Devnagari-Script & 46 & 1025 & 92000 & 1.00\\
\addlinespace
40927 & 167124 & CIFAR\_10 & 10 & 3073 & 60000 & 1.00\\
40966 & 146800 & MiceProtein & 8 & 82 & 1080 & 0.70\\
40975 & 146821 & car & 4 & 7 & 1728 & 0.05\\
40978 & 167125 & Internet-Advertisements & 2 & 1559 & 3279 & 0.16\\
40979 & 146824 & mfeat-pixel & 10 & 241 & 2000 & 1.00\\
\addlinespace
40982 & 146817 & steel-plates-fault & 7 & 28 & 1941 & 0.08\\
40983 & 146820 & wilt & 2 & 6 & 4839 & 0.06\\
40984 & 146822 & segment & 7 & 20 & 2310 & 1.00\\
40994 & 146819 & climate-model-simulation-crashes & 2 & 21 & 540 & 0.09\\
40996 & 146825 & Fashion-MNIST & 10 & 785 & 70000 & 1.00\\
\addlinespace
41027 & 167119 & jungle\_chess\_2pcs\_raw\_endgame\_complete & 3 & 7 & 44819 & 0.19\\

\bottomrule
\end{tabular}
\end{adjustbox}
\end{table}

\section{Useful links}
\label{sec:links}

We now collect all relevant links in a single place to simplify access to online material on OpenML benchmarking studies:
\begin{itemize}
    \item General online documentation: \url{https://docs.openml.org}
    \item Online documentation on benchmarking suites: \url{https://docs.openml.org/benchmark}
    \item Github repository with additional material, including a notebook to create updated suites: \url{https://github.com/openml/benchmark-suites}
    \item Github organization for OpenML.org: \url{https://github.com/openml}
    \item Python package: \href{https://pypi.org/project/openml/}{\emph{OpenML}} (PyPI)
    \item R package: \href{https://CRAN.R-project.org/package=OpenML}{\emph{OpenML}} (CRAN)
    \item Java package: \href{https://mvnrepository.com/artifact/org.openml/openmlweka}{\emph{org.openml.openmlweka}} (Maven Central)
\end{itemize}

\section{Suggested curation protocol}
\label{sec:protocol}

In this section we give an exemplary curation protocol for constructing new benchmarking suites. It is based on our experience constructing the OpenML-CC18 and its predecessor, the OpenML100. Steps can be removed or added depending on the desired benchmark purpose, the steps below serve as a guideline. 
\begin{enumerate}
    \item Steps that can be automated:
    \begin{enumerate}
        \item Specify the OpenML task type, for example supervised classification or supervised regression.
        \item Specify criteria on dataset properties, such as the size of the dataset, the number of features or the number of classes.
        \item Specify criteria on the data modalities that are supposed to be in the data. Currently, OpenML supports numerical, categorical, date and string.
        \item Specify whether the data should be sparse or not.
        \item Specify whether the data should contain missing values or not.
        \item Check whether tasks are too easy, either by querying for existing results on OpenML or by running machine learning algorithms locally.
    \end{enumerate}
    \item Steps that cannot be automated and should be performed on the outcome of the previous, automated steps. For our benchmark the following manual steps were added:
    \begin{enumerate}
        \item Check for artificial datasets.
        \item Check for dataset that require grouped or time-aware splitting.
        \item Check for datasets that are subsets of larger datasets (or binarized datasets in case of classification).
        \item Check for other forms of derived datasets, for example versions that do no longer contain feature names or only a subset of features.
        \item Check that all remaining datasets feature a reference.
    \end{enumerate}
\end{enumerate}

\fi

\end{document}